# What drives a goalkeepers' decision?

Research Track

Samer Fatayri, Kirill Serykh[1], and Egor Gumin[2]

1 Sportec Solutions AG, Unterföhring, Germany
2 Founder at football stats platform: xglab.pro

## Abstract

In soccer games, the goalkeeper's performance is an important factor to the success of the whole team. Despite the goalkeeper's importance, little attention has been paid to their performance in events and tracking data. Here, we developed a model to predict which movements would be most effective for shot-stopping and compare it to the real-life behavior of goalkeepers. This model evaluates the performance of goalkeepers based on their position and dive radius. We found that contrary to the movements that were considered most effective by our model, real-life goalkeepers' movements were more diverse. We further used our model to develop a tool to analyse goalkeepers' behavior in real-life soccer games. In addition, a simulator function allows team analysts or couches to identify situations that allow further improvement of the reaction of the goalkeeper.

Corresponding author: Samer Fatayri, fatayri@gmail.com

# Introduction

A goalkeeper is a crucial player in the defense of a soccer team, who is responsible for intercepting or stopping the ball before it crosses the final line and thus preventing the opposing team from scoring a goal. To prevent the attacking team from succeeding, the right positioning and defending movement of goalkeepers is of uttermost importance for a successful defense of the goal area.

With the growing popularity of event and tracking data for measuring and analyzing field players' performances and decision-making, goalkeepers are still severely underrepresented. The main metrics for tracking goalkeepers' performance do mostly focus on the percentage of shots stopped. This could be either measured in form of raw save percentage or with more advanced metrics such as postshot xG minus the number of goals conceded. In the following, we state the three problems with those metrics:

1. These metrics are descriptive and not prescriptive. Using these metrics, we can compare goalkeepers' performances with each other, however, these metrics do not provide any information on how to improve the performance of goalkeepers.
2. A poor repeatability of goalkeepers' performances. The performance of future seasons is not necessarily related to the performance of past seasons (Antonio, 2014). Hence, it is difficult to scout goalkeepers using data-driven approaches.
3. A small sample size. On average, a goalkeeper faces 4.5 shots on target per game, which equates to approximately 150 shots per season. This results in large standard deviations and a large number of outliers.

The aim of this research is to investigate how goalkeepers can prevent a shoot from crossing the final line. Therefore, we consider different aspects of the defense process. The process of shot-stopping can be categorized into different components, including the positioning of the goalkeeper, reaction quality, general technique, etc. The aim of this research is to develop a framework to evaluate a goalkeeper's positioning and to be able to advise on how to improve the goalkeepers' performance in different situations.

# Framework design

**2.1 Simulated Shots and Positions**

The main goal of this framework is to evaluate the quality of the goalkeeper's decision-making during the attack of the opponents. Here, we compare the decision the goalkeeper made to the potential decisions the goalkeeper could have made.

We combined all the events that arose when defending 30% of the pitch toward the goal area, and when the ball was controlled by the opponent's team. When these conditions were met, we simulated a shot being made toward the goal. Our main interest was to investigate whether the goalkeeper's current position would be optimal to prevent a shot from entering the goal or whether the goalkeeper's position could be improved. Therefore, we simulated shots and applied them to different, simulated positions of the goalkeeper.

For simulating the goalkeeper's position, we introduced a simple "run model", which takes the time difference between the current and the previous events and returns the radius the goalkeeper could have covered by running (Figure 1). Adding this radius to the goalkeeper's location (blue) from the previous event with 8 angles (0, π/4, π/2, π, …, 7/4 π), we get 8 points to which the goalkeeper could have run (red). This run model is very simplistic and does not cover all potential scenarios (an example of simulated positions is shown in figure 1).

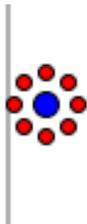

Figure 1: Run model. The run model visualizes where the goalkeeper (current position in blue) could have run(red). The grey lines mark the goal area, the double grey line the goal line.

For simulating shots onto the goal, we chose 6 different points in the goal area (Figure 2). Two important points on how shots were simulated:
1. We considered that every shot reached the target and would exactly land at one of the 6 points (orange). In this scenario, the goalkeeper was challenged with strikers performing a perfectly accurate shot.
3

2. We assumed that the striker would choose one of the 6 points with an equal probability, independent of the striker's position on the pitch.

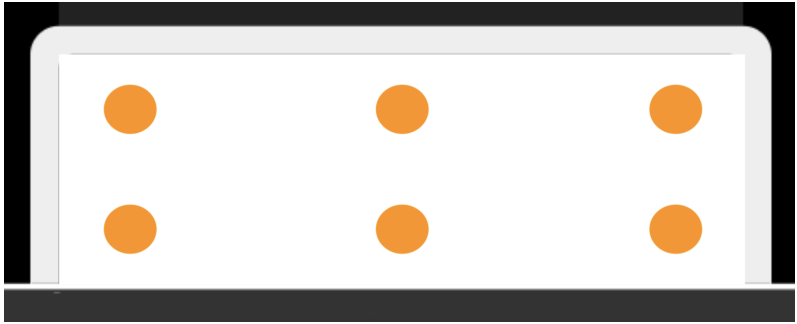

Figure 2: Simulated shot end locations in goal. Six potential points where the striker's shot will end up in the goal (orange). The goal area is depicted in black.

## 2.2 Probability of Goal

As all the simulated shots reach the target, we defined the probability of a goal for a single shot by its end location as:

$$P(goal) = P(not\ blocked) * P(not\ saved | not\ blocked)$$

## 2.3 Goalkeeper's position metric

We define the probability of a goal in relation to the goalkeeper's position, as the maximum of the probabilities for all 6 simulated positions:

$$P(goal | position) = \max_{shots} P(goal | shot)$$

# Save and block models

### 3.1 Block model
To develop a block model we used two sets of features:
1. The density of players between the shooter and the goal.
2. The time that the player will need to get to the trajectory of the shot (all shots are modeled by straight lines).

Using these features, we developed an XGBoost classifier that predicts the probability of a shot being blocked by the field players.

### 3.2 Save model

Using a save model, we introduced different types of shadows that describe the goalkeeper's position at a particular moment of time.

### 3.2.2 Position shadow

Based on the location of the goalkeeper and the shooter, we defined 2 different triangles (Figure 3):
1. Triangle ABC depict the projection of the goalkeeper's position at the goal (vertex A - position of goalkeeper, B - right post, C - left post)
2. Triangle OBC depict the projection of where the shooter can make a shot (vertex O - shooter position)

Thus, we defined position shadows as the ratio of the area of the intersection between these 2 triangles to the area of the second triangle.

$$Position\ Shadow\ =\ \frac{S(ABC \cap OBC)}{S(OBC)}$$

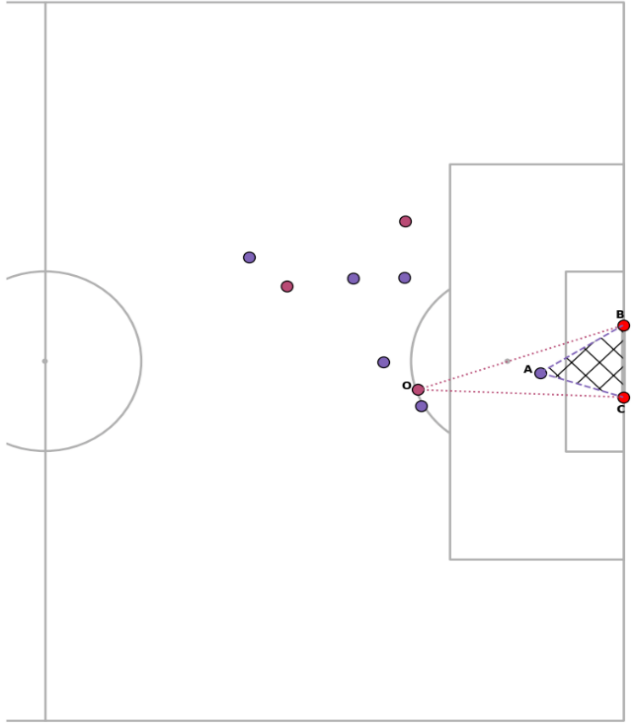

Figure 3: Position shadows. Triangle ABC shows the projection of the goalkeeper at the goal. Triangle OBC shows the projection of the shooter and the potential shoots onto the goal.

### 3.2.3 Goal Shadow

The goal shadow displays the ratio of the goal area covered by the potential dive area of the goalkeeper. We are employing the dive model of Ibrahim et.al (2018), using the following constants: reaction time = 0.2 seconds, jump time = 0.5 seconds, maximum dive time = 1.2 seconds. For simplicity, we are approximating the goalkeeper's dive area as a rectangle, where the maximum height of the jump equates to the goalkeeper's height + 50 cm.

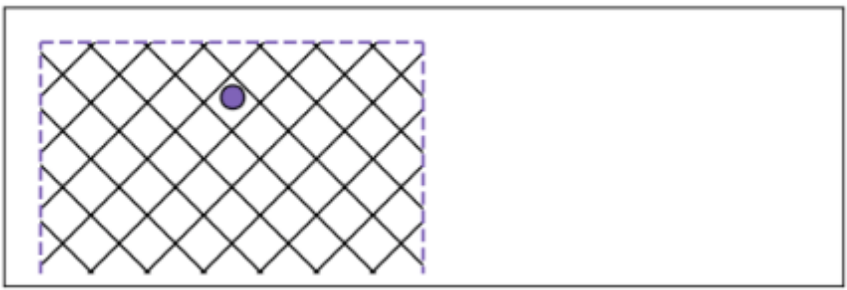

Figure 4: Goal shadow. The goalkeeper's (purple dot) dive area (cross pattern) in the goal area (black rectangle).

### 3.2.4 Dive shadow

To define the third shadow, we first calculate the area of the potential surface, which the goalkeeper could reach by hand in case of a dive (Figure 5). We defined this surface as a circle with a radius equal to the orthogonal to the line of the shot, assuming that the goalkeeper knows where the shot would take place. In figure 5, D is the point where the line of the shot intersects the orthogonal of the goalkeeper position, S. S is the end location of one of 6 simulated shots, to be precise, the left bottom corner. The orthogonal might reflect a much longer distance than the goalkeeper could cover within a dive, hence, we used the circle displaying the maximum distance covered by the goalkeeper's dive in a period of 1.2 seconds.

$$Dive\ Shadow = \frac{S(Circle(AD) \cap OBC)}{S(OBC)}$$

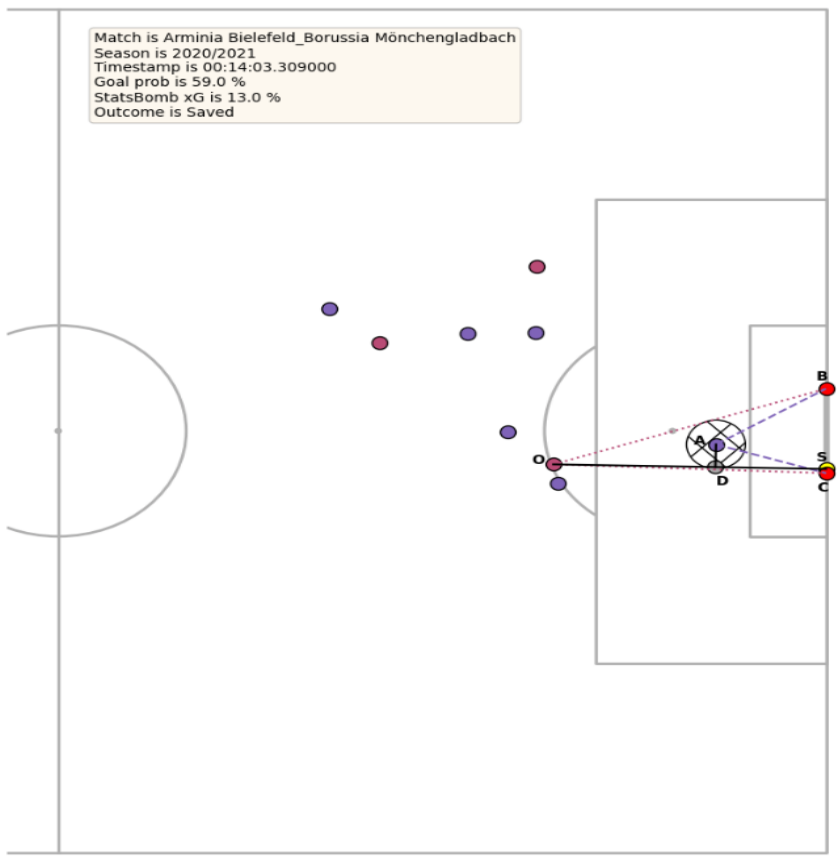

Figure 5: Goalkeeper's dive shadow. The dive shadow is the potential area (cross patterns) the goalkeeper could reach with a dive.

### 3.3 Prediction model

We continued to further add features to the above shadows to improve our prediction model. Among others, we added the distance of the shot, the angles of the shot to the goal between the shooter and the goalkeeper, and the shooter under pressure. Using these features, we trained an XGBoost classifier that predicts the probability that the goalkeeper will save the shot, if the shot arrives at the goal.

# Results

Figure 6 displays the distribution of all moves that our model considers to be optimal for the goalkeeper to prevent a shot to enter the goal area. The thick black line represents the goal line. The model found backward-going patterns to be more effective in preventing a goal than forward going patterns.

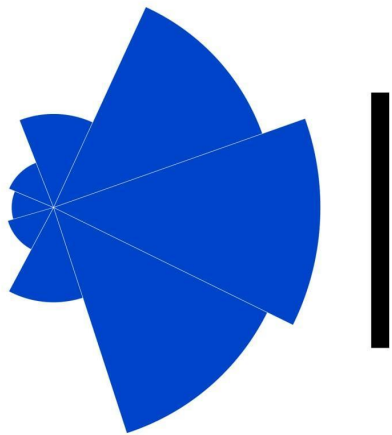

Figure 6: Model of goalkeeper's moves that prevent a shot from entering the goal. The black line represents the goal line. Blue areas depict the movement area of the goalkeeper. The larger the blue area the higher the probability of shoot-stopping.

In the following, we show how the model's behavior compares to real-life goalkeepers' decisions (Figure 7). We found that the distribution of moves differs to our model predicted behavior (Figure 6). In real-life situations, goalkeepers tend to make much less conservative decisions, by protecting the goal instead of attacking the ball and move more frequently toward the ball or to the sides of the goal area instead of backward as found favorable by our model.

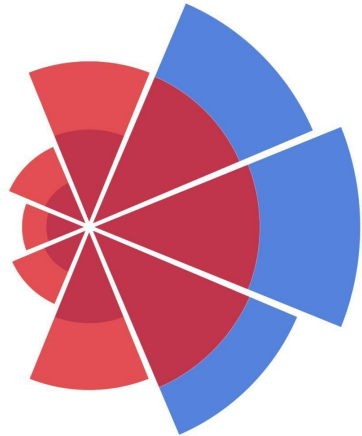

Figure 7: Comparison of real-life decision-making of the goalkeeper (red) with the models' predicted preferential decisions (blue).

## A Tool to analyse a goalkeeper's performance

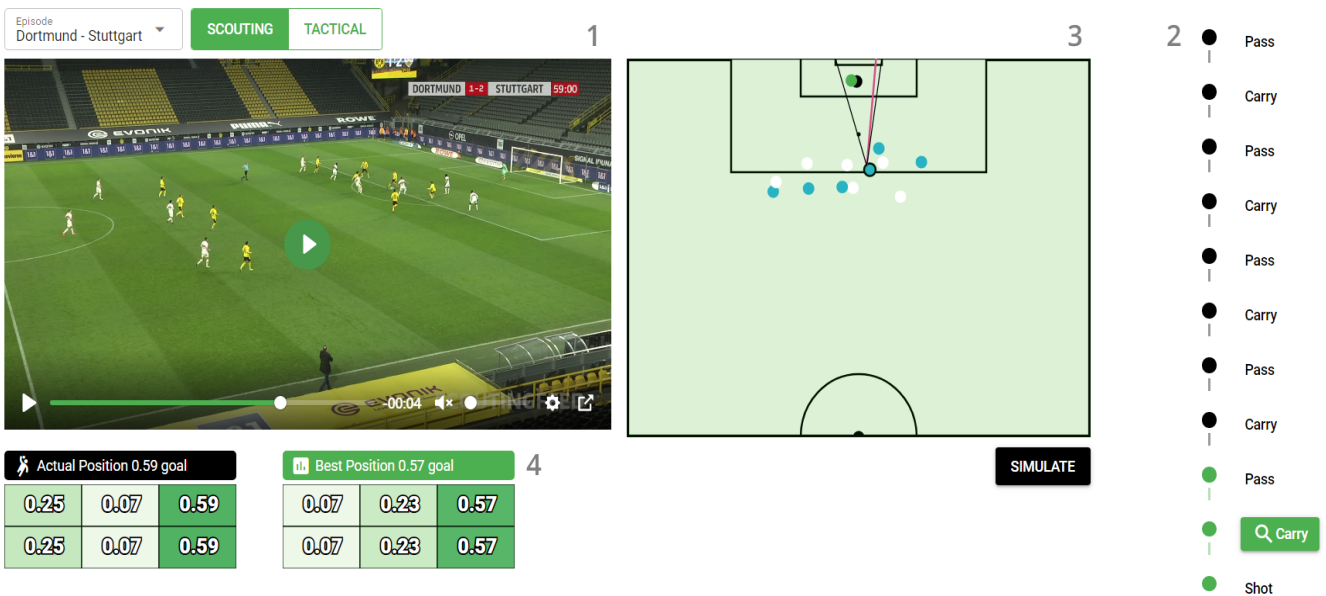

Figure 8: Tool to analyse a goalkeeper's performance. The tool can be used to select specific episodes of the game and analyse whether the goalkeeper's position could be improved. (1) Video player to select specific episodes for analysis (2) Episode list: Green highlighted events allow the position model to be applied. The position model cannot be applied to black-highlighted events. (3) The field view displays a real-time update of the players in the video (4) Goal view shows the probability of the goalkeeper to shshot-stop. The darker green, the lower the probability.

We developed a tool that can be used by team analysts or goalkeeper coaches to identify situations where a goalkeeper's position could be potentially improved (Figure 8).
Games were split into episodes. An episode was defined as a sequence of events where the last event is a shot toward the goal. An episode has an approximate length of 10-15 seconds.

### 5.1.1 Video Player
To analyse a game, analysts can select an episode and play the video, rewind it and also change the playback speed. Tactical and scouting feeds are available.

### 5.1.2 Episodes list
The episode list shows which event is currently happening in the episode of the game. This feature can be used to rewind the video and to analyse a specific event. The positioning model can be applied to all green-highlighted events. The remaining events are highlighted in black and refer to events in which the position of the goalkeeper is unknown.

### 5.1.3 Field View
While the video is playing, players' positions from the freeze frames are synchronized and updated on a field view in real-time. Players of the defending team are shown as white pins, whereas players of the attacking team are depicted in blue. Blue pins with black borders reflect players of the attacking team who perform an attack on the goal. The goalkeeper's actual position is marked with a black pin. Green pins reveal the model's suggested positions for the goalkeeper to improve the defense of the goal. The red line shows the angle which would lead to the most successful shoot by the attacking player (blue pin with black borders).

### 5.1.4 Goal projection

The goal projection is a front view of the goal, which is split into cells showing the calculated probability of conceding a goal for each cell. Dark-colored cells are the hardest to defend for a goalkeeper in a specific position.

## 5.2 Simulator

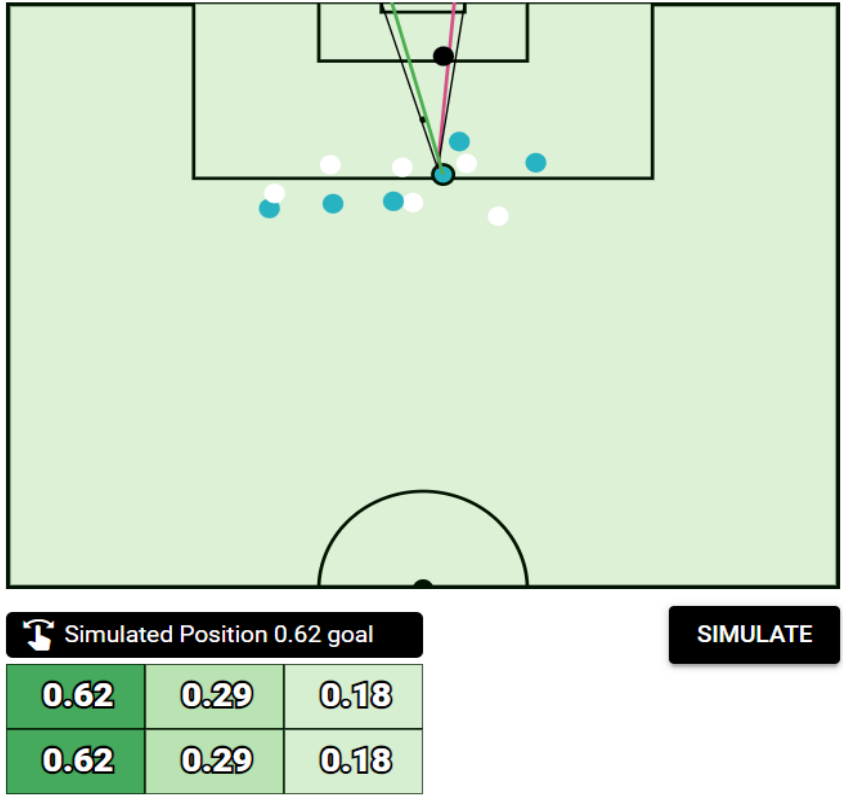

Figure 9: Simulator to evaluate goalkeeper's performance in different positions. The red line points to the unprotected part of the goal for the goalkeeper's current position. The green line indicates the least protected point of the goal for the goalkeeper's simulated position (black pin).

The simulator is used to gauge the goalkeeper's performance in different positions. Analysts can move any player on the interactive field view to simulate different players' behavior options. Furthermore, the goalkeeper's position, defenders' locations or an attacking player with a ball can be moved arbitrarily. The red line drawn by the simulator points to the unprotected part of the goal considering the goalkeeper's current position The green line indicates the least protected point of the goal for the goalkeeper's simulated position (black pin).

## Discussion

Using save and block models, we developed a tool to analyse the performance of goalkeepers and a simulator to evaluate how goalkeepers could improve their performance.

Firstly, we compared the results of our predictive model with the real-life behavior of goalkeepers. While the model found backward movement towards the goal line to be more effective to block the ball from entering the goal, real-life goalkeepers tend to move forward, away from the goal line, in the direction of the ball. We assume that the main reason for such a big difference is the assumption about the equal probabilities of the shot end locations. The simulated goalkeepers tend to be minimizing the maximum probability of goals overall, while real-life goalkeepers have an intuition to which side of the goal the striker is going to shoot.

Secondly, we developed a tool to analyse goalkeepers' behavior in real-life soccer games. This tool can be useful for team analysts or couches to identify situations that allow further improvement of the reaction of the goalkeeper. Moreover, the tool allows to analyse a fictive game setting. Players can be moved arbitrarily to evaluate different scenarios.

We believe that this tool can be valuable in improving goalkeepers' performance. Nevertheless, our model could be further improved by including an estimate of off-target shoots, improving our dive model to estimate the length of the jump more accurately, taking the anthropometry of the goalkeeper into account, expanding the functionality of our tool by adding more screen views and more ways to customize the simulation.

## Conclusion

We defined a new framework to objectively assess a goalkeeper's shot-stopping. StatsBomb 360 data is adding more context that affects the decisions of the goalkeeper, and we believe that within this framework it is possible to get a more tedious quantitative estimate of goalkeepers' performance.

## Acknowledgements

We would like to thank Francisco Goitia (Lead Machine Learning Engineer @ StatsBomb) for mentoring our team during the time of preparation for the StatsBomb Conference in London, UK, 2021.